\documentclass[10pt,twocolumn,letterpaper]{article}

\usepackage{cvpr}              

\makeatletter
\@namedef{ver@everyshi.sty}{}
\makeatother
\usepackage{graphicx}
\usepackage{amsmath}
\usepackage{amssymb}
\usepackage{booktabs}
\usepackage{utfsym}
\usepackage{multirow}

\usepackage[pagebackref,breaklinks,colorlinks]{hyperref}

\usepackage[capitalize]{cleveref}
\crefname{section}{Sec.}{Secs.}
\Crefname{section}{Section}{Sections}
\Crefname{table}{Table}{Tables}
\crefname{table}{Tab.}{Tabs.}

\def\cvprPaperID{88} 
\def\confName{CVPRW}
\def\confYear{2023}

\begin{document}

\title{DIPNet: Efficiency Distillation and Iterative Pruning for Image Super-Resolution}



\author{Lei Yu\,$^{1*}$\quad
Xinpeng Li\,$^1$\thanks{Equal contribution}\quad
Youwei Li\,$^2$\quad
Ting Jiang\,$^1$\quad
Qi Wu\,$^1$\quad
Haoqiang Fan\,$^1$\quad
Shuaicheng Liu\,$^{3,1}$\thanks{Corresponding author}  \\
$^1$\,Megvii Technology\quad$^2$\,Microbt\\
$^3$\,University of Electronic Science and Technology of China\\
\tt\small \{yulei02, lixinpeng\}@megvii.com, liyouwei.wellee@gmail.com, \\
\tt\small \{jianting, wuqi02, fhq\}@megvii.com, liushuaicheng@uestc.edu.cn
}
\maketitle

\begin{abstract}
Efficient deep learning-based approaches have achieved remarkable performance in single image super-resolution. However, recent studies on efficient super-resolution have mainly focused on reducing the number of parameters and floating-point operations through various network designs. Although these methods can decrease the number of parameters and floating-point operations, they may not necessarily reduce actual running time. To address this issue, we propose a novel multi-stage lightweight network boosting method, which can enable lightweight networks to achieve outstanding performance. Specifically, we leverage enhanced high-resolution output as additional supervision to improve the learning ability of lightweight student networks. Upon convergence of the student network, we further simplify our network structure to a more lightweight level using reparameterization techniques and iterative network pruning. Meanwhile, we adopt an effective lightweight network training strategy that combines multi-anchor distillation and progressive learning, enabling the lightweight network to achieve outstanding performance. Ultimately, our proposed method achieves the fastest inference time among all participants in the NTIRE 2023 efficient super-resolution challenge while maintaining competitive super-resolution performance. Additionally, extensive experiments are conducted to demonstrate the effectiveness of the proposed components. The results show that our approach achieves comparable performance in representative dataset DIV2K, both qualitatively and quantitatively, with faster inference and fewer number of network parameters.
\end{abstract}

\section{Introduction}
\label{sec:intro}
\begin{figure}[t]
\centering
\includegraphics[width=0.95\linewidth]{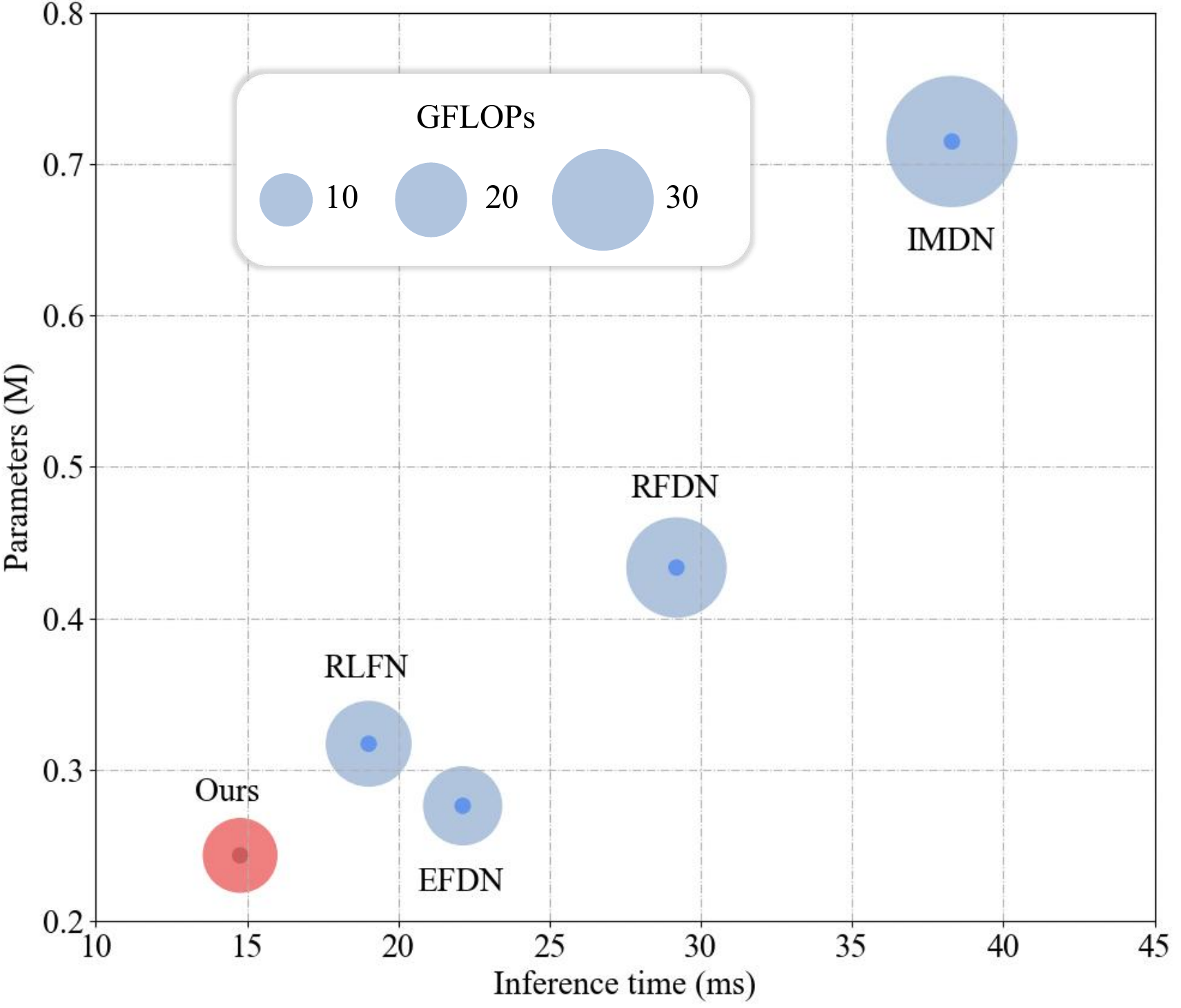}
\caption{Comparison with recent efficient SR methods. The figure shows the GFLOPs of these methods when the input is 256\(\times\)256, the number of parameters of these models and their average inference time using NVIDIA 2080ti under the DIV2K validation set.} \label{fig:teaser}
\end{figure}

Single Image Super-Resolution (SISR) aims to reconstruct a high-resolution~(HR) image from a low-resolution~(LR) input and has become an essential task in low-level computer vision for enhancing image resolution\cite{chen2018fsrnet,fu2019hyperspectral}. Recent SISR approaches\cite{chen2018fsrnet,dong2014learning,hui2019lightweight,ignatov2021real,tai2017image,shi2016real,liu2020residual,wei2020aim,zhang2021edge} based on deep learning have achieved great success by significantly improving the quality of reconstructed images. However, these methods frequently require large amounts of computational resources, making it challenging to deploy them on resource-constrained devices for real-world applications.

To address this issue, there is a growing need to develop efficient SISR models with higher inference speed while maintaining good trade-offs between image quality and computation cost. Prior research has attempted to reduce model parameters or floating-point operations (FLOPs) to improve efficiency. Recursive networks with weight-sharing strategies are often used to decrease the number of parameters, but they may not necessarily reduce the number of operations and inference time due to their complex graph topology. Similarly, commonly used techniques to reduce FLOPs, such as depth-wise convolutions, feature splitting, and shuffling\cite{ahn2018fast,hui2019lightweight,hui2018fast,liu2020residual}, may not always improve computational efficiency.

Therefore, we consider the problem of achieving efficient super-resolution from another perspective, that is, how to obtain an efficient super-resolution model only through better training strategies without too much additional network design. As shown in Fig.~\ref{fig:teaser}, the strategies we proposed make the SR model faster and smaller. It is difficult to train a small network directly, and it is easier to train a large teacher network first and then guide the small student network to learn through knowledge distillation. However, due to the large gap between the learning ability of the teacher network and the student network, it is difficult for the student network to learn enough high-frequency information if it is directly distilled. Inspired by HGGT~\cite{chen2023human}, the HR images we used are not the original HR images in the data set, but the enhanced HR images. The enhanced images can provide richer high-frequency information, which can help student network with limited learning ability to learn more easily.
Inspired by repVGG\cite{ding2021repvgg}, we designed the network structure including series branches, parallel branches and residuals in the process of designing the student network. These additional branches can increase the learning ability of the student network. When the student network training converges, it can be reparameterized and simplified into a lightweight structure. This operation enhances the learning ability of the model without introducing additional model complexity. In order to make full use of the guiding ability of the teacher network, we use a multi-level distillation strategy, that is, set anchor points at different nodes of the network, and use the features of different levels at the anchor points to perform distillation.

Usually, for the convenience of training in image tasks, we will use relatively small patches for training. However, recent studies\cite{chu2022improving} pointed out that using this method will cause the input distribution to be different during training and testing so that it cannot be effective enough when the network performs some global operations during testing. Therefore, in order to make the model perform better in the testing phase, the input size during training and testing should be as close as possible. However, directly using large-sized patches for training will make the training very time-consuming, and the inability to use larger batches will further affect the stability of training. At the same time, it is not conducive to using large patches as input for networks with limited expressive capabilities to mine global information. So we adopted a progressive learning method, that is, gradually increasing the input patch during training, and achieved good results. The trained student network still has some unimportant redundant parameters, so we further iteratively pruned the model to further compress the model size.

Our contributions can be summarized as follows:
\begin{itemize}
\item For the first time, we propose the use of enhanced HR images to improve the learning ability of lightweight networks.
\item We propose a novel multi-stage lightweight training strategy combining distillation, progressive learning, and pruning.
\item We conducted extensive experiments to demonstrate the effectiveness of our method, and our method outperformed all other competitors in the NTIRE 2023 efficient super-resolution challenge in terms of time consumption and model size.
\end{itemize}
\section{Related Work}

\subsection{Single Image Super-Resolution}

In the past few years, deep neural networks (DNN) have shown remarkable capability on improving SISR performance. The pioneering work is SRCNN~\cite{simonyan2014very} which applies the bicubic downsampling on HR images to construct data pairs and employs a simple convolution neural networks (CNN) to learn the end-to-end mapping from LR to HR images. Then plenty of CNN-based methods have been proposed to achieve better performance~\cite{kim2016accurate,lim2017enhanced,shi2016real,zhang2018residual,haris2018deep,tai2017image,luo2021ebsr,zhang2018image,zhang2020residual,luo2022bsrt}. For example, Kim {\textit{et} \textit{al}.}~\cite{kim2016accurate} proposed a 20-layer network with residual learning, which inspired the development of deeper and wider networks for SISR. EDSR~\cite{lim2017enhanced} followed the idea of residual learning and modified residual blocks by removing the batch normalization layer to build a very deep and wide network. RCAN~\cite{zhang2018image} have introduced channel attention and second-order channel attention respectively, which exploit feature correlations for improved performance. Moreover, recent works~\cite{ledig2017photo,sajjadi2017enhancenet,wang2018esrgan,Ji_2020_CVPR_Workshops,li2022d2c} have been proposed to improve the perceptual visual quality of real-world images. Meanwhile, some studies based on blind image super-resolution were proposed~\cite{gu2019blind,bell2019blind,luo2022deep}, addressing the problem of degenerate kernels present in real-world super-resolution. In addition, some works employ some advanced losses such as the VGG loss~\cite{simonyan2014very}, perceptual loss~\cite{johnson2016perceptual}, and GAN loss~\cite{goodfellow2014generative} to learn realistic image details. Recently, transformer-based super-resolution methods~\cite{liang2021swinir, chen2022activating} have gained popularity, which achieve high performance. However, most of these methods require a large amount of computational resources and have a high number of parameters, FLOPs and inference time, and do not facilitate practical deployment and application in edge devices.

\begin{figure*}[t]
\centering
\includegraphics[width=0.95\linewidth]{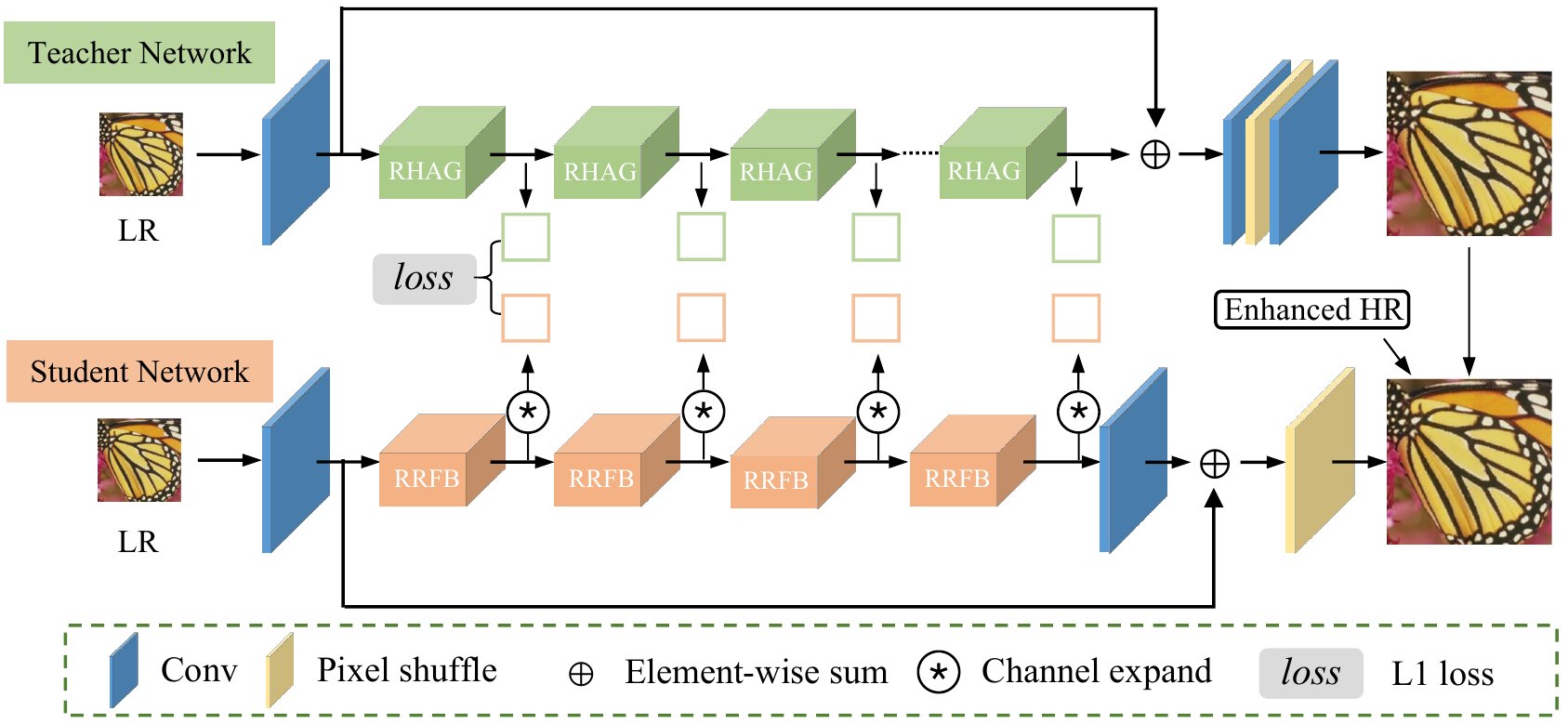}
\caption{The multi-stage feature distillation pipeline of our method. } \label{rep_megSR}
\end{figure*}

\subsection{Efficient Image Super-Resolution}
Efficient Image Super-Resolution aims to reduce the computational effort and the number of parameters of the SR network while achieving faster inference times and maintaining high performance. In real-world SR model deployments, the computing power of the deployed devices is often limited, such as edge devices, etc. In this case, the efficiency of the SR network becomes an important aspect.
To fit the increasing demands for deploying SR models with limited computing resources, numerous works have refocused their attention to efficient image SR techniques~\cite{ahn2018fast,hui2019lightweight,liu2020residual,kong2022residual,du2022fast,luo2023fast}. At the same time, a number of competitions, just like NTIRE and AIM, have launched efficient image SR entries to promote the development of relevant research~\cite{li2022ntire,ignatov2023efficient}. In recent related studies, CARN~\cite{ahn2018fast} presented local and global cascading mechanisms to achieve a lightweight SR network. IMDN~\cite{hui2019lightweight} designs an information multi-distillation network by constructing the cascaded information multi-distillation blocks to extract hierarchical features. The following work RFDN~\cite{liu2020residual} further improves the network by introducing feature distillation blocks that employ 1$\times$1 convolution layers to implement dimensional change. Based on RFDN, RLFN~\cite{kong2022residual} investigates its speed bottleneck and enhances its speed by removing the hierarchical distillation connections. Furthermore, RLFN proposes a feature extractor to extract more information of edges and textures. With these advancements, they achieved first place in the NTIRE 2022 Efficient Super-Resolution Challenge~\cite{li2022ntire}.

\section{Method}

We propose an efficiency distillation and iterative pruning SR network named DIPNet which consists of four main components. In Sec.~\ref{sec:na}, we revisit the RLFB and propose a reparameterization residual feature block (RRFB), and our network structure is mainly constructed by stacking multiple RRFBs, as shown in Fig.~\ref{rep_megSR}. In Sec.~\ref{sec:mggte}, we introduce the method of model-guided ground-truth enhancement strategy to improve the quality of original HR. Then we discuss the multi-anchor feature distillation in Sec.~\ref{sec:mfd}, which can effectively improve the performance of the network. Finally, we propose an iterative pruning strategy in Sec.~\ref{sec:ips} to further reduce the number of model parameters.

\subsection{Reparameterization Residual Feature Block}\label{sec:na}
\begin{figure}[t]
\centering
\includegraphics[width=1.0\linewidth]{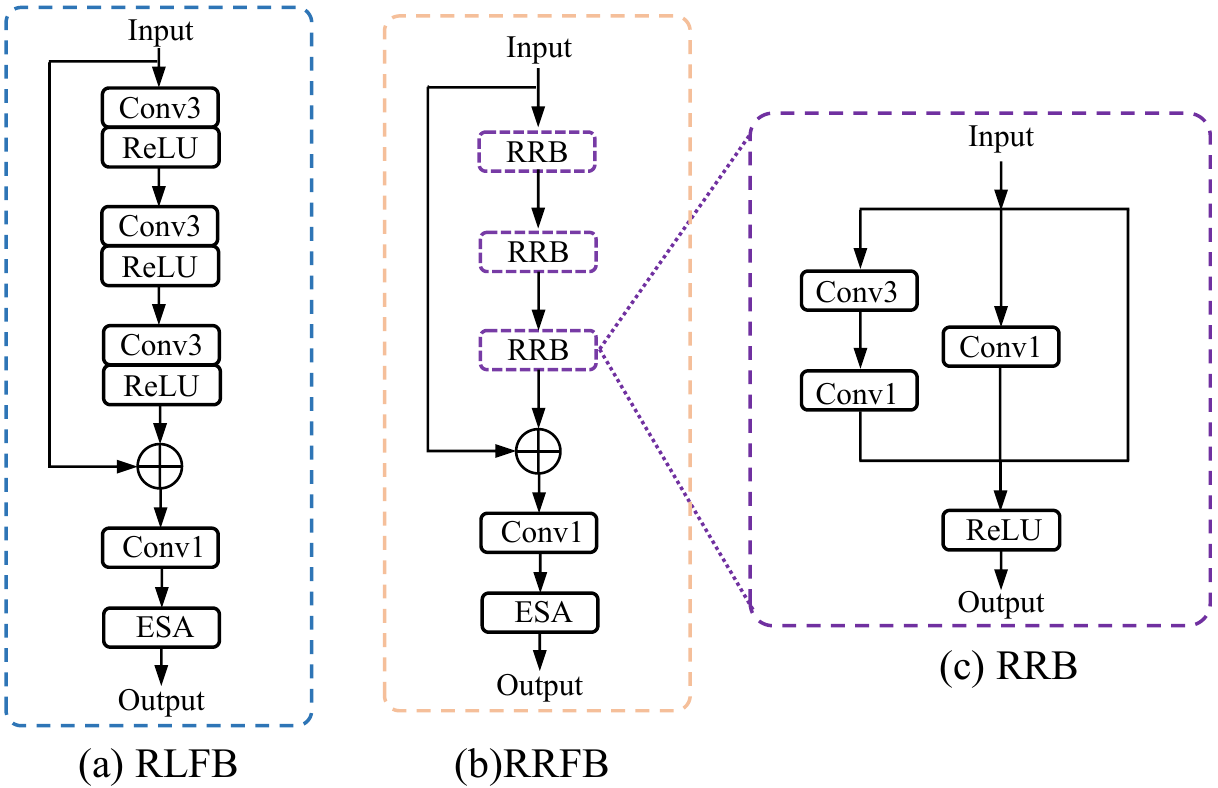}
\caption{ (a) Structure of RLFB. (b) The structure of  RRFB. } \label{rep}
\end{figure}

Following RFDN~\cite{liu2020residual} and RLFN~\cite{kong2022residual}, we also use an information distillation network to reconstruct high-quality SR images. Based on the block RLFB of RLFN, we introduce the re-parameterizable topology to the block. The original block of RLFB is shown in Fig.~\ref{rep}(a), we expand the RLFB in RLFN to the structure reparameterization residual feature block~(RRFB) shown in Fig.~\ref{rep}(b) in the training phase. The structure of RRB which is shown in Fig.~\ref{rep}(c) excavates the potential ability of complex structure during optimization, while maintaining computational efficiency, as it is computationally equivalent to a single 3x3 convolution during inference.

\subsection{Model Guided Ground-truth Enhancement}\label{sec:mggte}
\begin{figure}[h]
\centering
\includegraphics[width=0.98\linewidth]{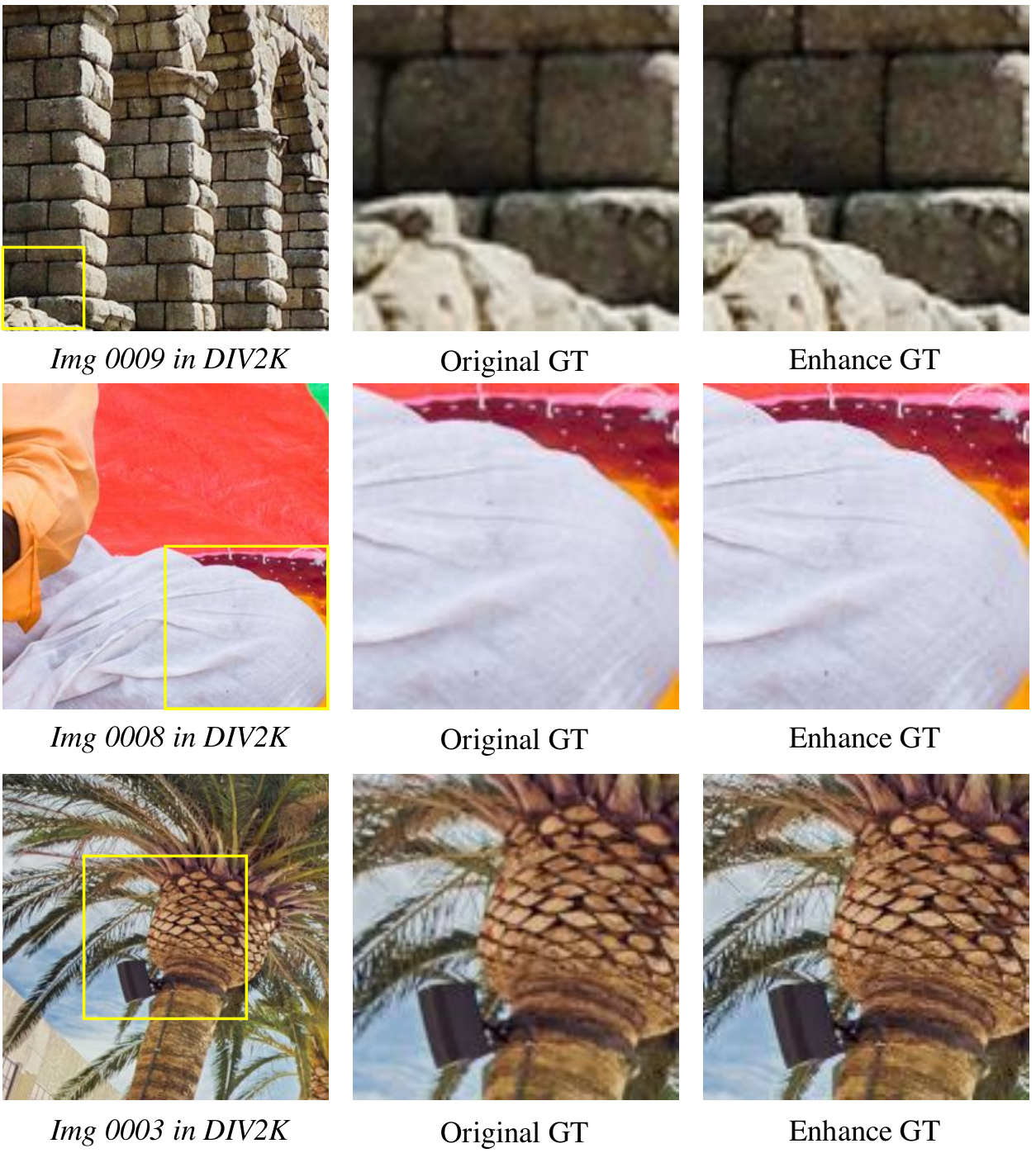}
\caption{ Visual comparison of the original GT and the enhance GT.} \label{fig:enhanceGT}
\end{figure}

According to our understanding, almost all existing SR methods directly use the original HR images in the training phase. However, the perceptual quality of the original HR images may not be high enough as mentioned by HGGT~\cite{chen2023human}. Inspired by HGGT, we proposed a model guided ground-truth (GT) enhancement strategy to enhance the quality of HR. We first train a large network with Hybrid Attention 
Transformer~\cite{chen2022activating} backbone for 1x super-resolution. The HR image \(I_{HR}\) is then utilized as input for 1x super-resolution, yielding an enhanced HR output \(I_{enh}\). Then the low-resolution image \(I_{LR}\) and the enhanced HR \(I_{enh}\) is used for 4x super-resolution training. Different from HGGT, we do not conduct manual patch selection and retain the patch in the flat area since that manual selection tends to generate results that were more visually pleasing, but may not improve objective metrics.  As shown in Fig.~\ref{fig:enhanceGT}, the quality of the enhanced ground-truth in some patches is significantly better than the original ground-truth.

\subsection{Multi-anchor Feature Distillation}\label{sec:mfd}
In order to further enhance the performance of our lightweight model, we proposed a multi-anchor feature distillation method. As illustrated in Fig.~\ref{rep_megSR}, our multi-anchor feature distillation consists of two stages. In the first stage, we also train a large teacher network HAT~\cite{chen2022activating}, denoted as \(\mathcal{T}\). It is worth noting that we use the enhance high-resolution image \(I_{enh}\) as discussed in Sec.~\ref{sec:mggte} for 4x super-resolution training through minimizing the following loss:
\begin{equation}
    L_{\mathcal{T}}= ||\mathcal{T}(I_{LR})-I_{enh}||_1.
\end{equation}
After training the teacher network, we perform a multi-level distillation on the proposed student network, denoted as \(\mathcal{S}\). Once the student network training converges, it can be restored to the RLFB structure by the reparameterization technique. During distillation, we use the feature maps extracted from four different depths of \(\mathcal{T}\) to supervise the learning of each of the four blocks in \(\mathcal{S}\). Specifically, we minimize the following losses:
\begin{equation}
L_{feat}=\lambda_i \sum_{i=1}^{4}  ||F^\mathcal{T}_{i}- \psi(F^\mathcal{S}_i)||_1,
\end{equation}
where \(F^\mathcal{S}_i\) represents the feature map of the output of the \(i\)-th block of  \(\mathcal{S}\), while \(F^\mathcal{T}\) represents the feature of the output of some residual hybrid attention groups~(RHAGs) of \(\mathcal{T}\), \(\psi\) represents the operation of using a \(1\times1\) convolution to expand the feature channels of \(\mathcal{S}\) to the number of feature channels of 
\(\mathcal{T}\), and \(\lambda_i\) is a weight for controlling the importance of the supervision from each depth level.

We also use the outputs of \(\mathcal{T}\) as pseudo ground-truth and the enhanced ground-truth to further supervise the learning of \(\mathcal{S}\). Specifically, we compute the following losses:
\begin{equation}
L_{out}= ||\mathcal{T}(I_{LR})-\mathcal{S}(I_{LR})||_1+||\mathcal{S}(I_{LR})-I_{enh}||_1,
\end{equation}
During distillation, the final loss is a combination of \(L_{feat}\) and \(L_{out}\):
\begin{equation}
L_{dis}= L_{feat}+L_{out},
\end{equation}
After distillation, we employ a progressive learning strategy to finetune \(\mathcal{S}\). We gradually increase the size of the input patch while using \(L_2\) loss for supervised training until the model fully converges:
\begin{equation}
L_{pl}= ||\mathcal{S}(I_{LR})-I_{enh}||_2^2,
\end{equation}

\begin{figure*}[t]
\centering
\includegraphics[width=0.9\linewidth]{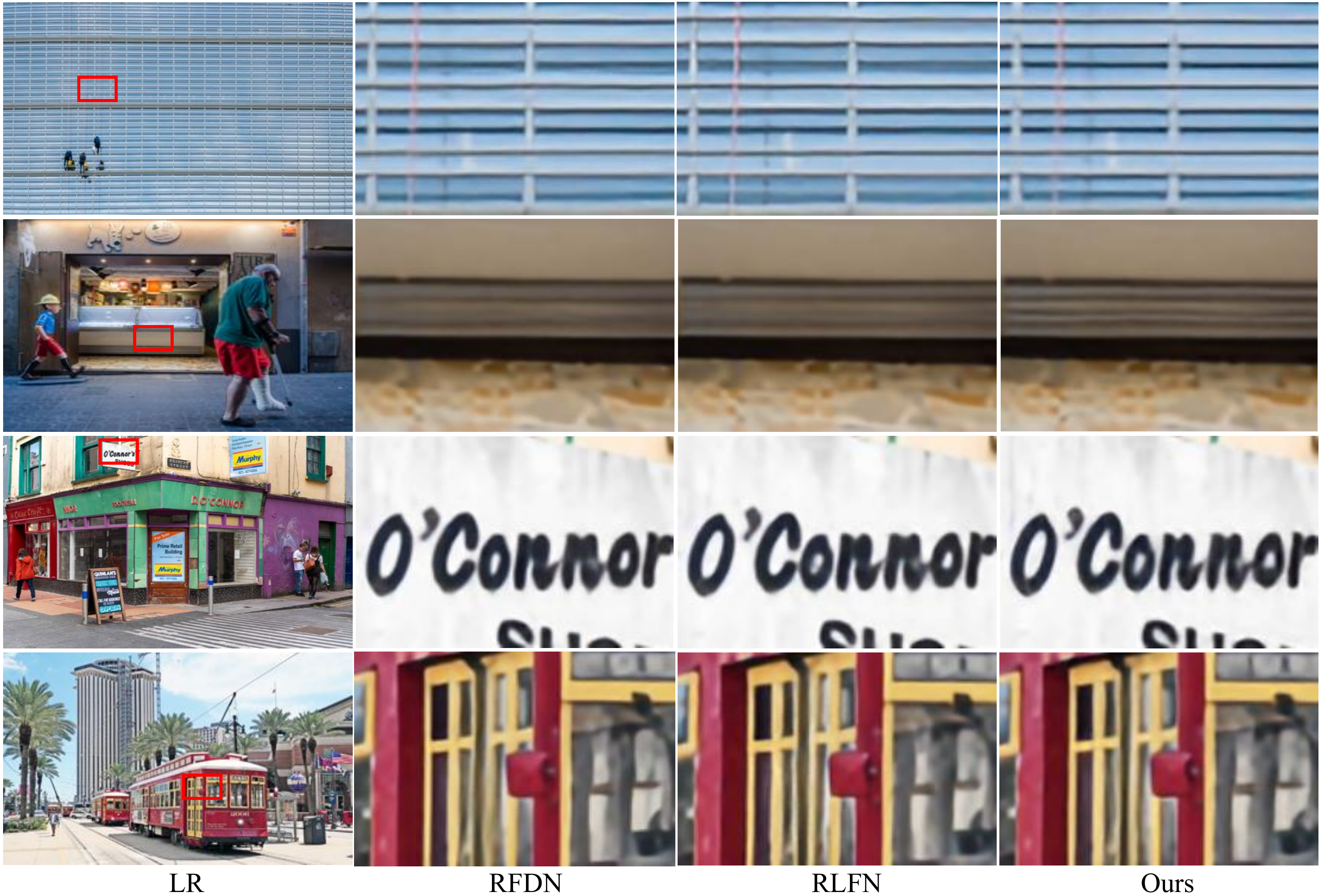}
\caption{Visual comparison of the results of ours and other methods on the validation set of DIV2K.} \label{fig:resvis}
\end{figure*}
\subsection{Iterative Pruning Strategy}\label{sec:ips}
Finally, we iteratively pruned the reparameterized student network \(\mathcal{S}\):
\begin{equation}
\begin{aligned}
\mathcal{S}_p^{i} =\varphi (\Phi(\mathcal{S}_p^{i-1};r)),\\
\end{aligned}
\end{equation}
where \(\Phi\) is the pruning operation, \(r\) is the pruning rate, \(\varphi\) means the finetuning operation, \(\mathcal{S}_p^{i}\) means the network after the \(i\)-th pruning. Inspired by AGP~\cite{zhu2017prune}, \(L_2\) filter pruning is used in our iterative pruning method for model training. We stop pruning until the network cannot make effective predictions, and use the network obtained from the last effective pruning as the final network.

\begin{figure*}[t]
\centering
\includegraphics[width=0.85 \linewidth]{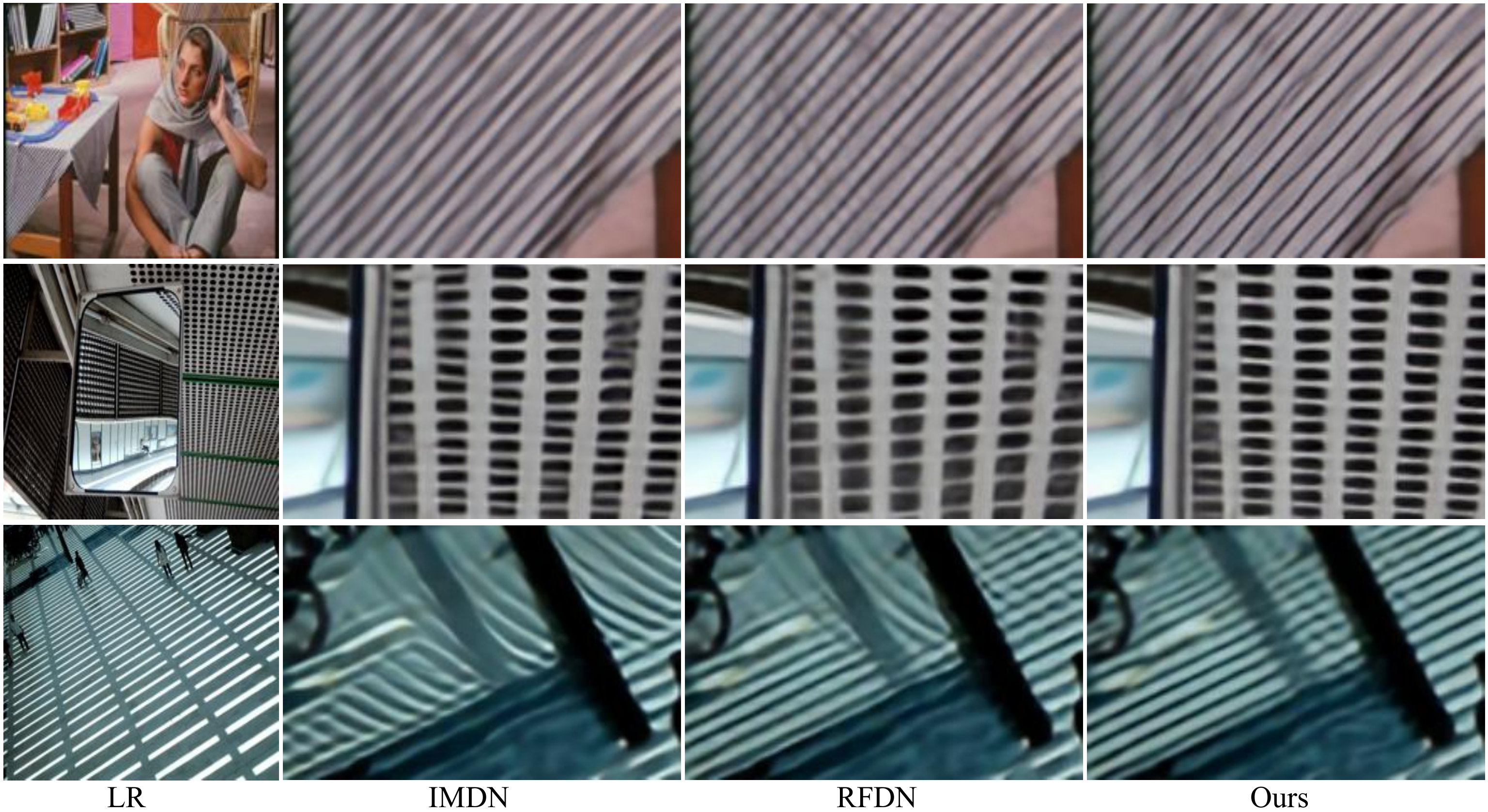}
\caption{More visual comparison on the other benchmarks.} \label{fig:vis2}
\end{figure*}
\begin{table*}[t]
\centering
\setlength{\tabcolsep}{2.9mm}{
\begin{tabular}{ccccccc}
\hline
Method & PSNR  & Time (ms) & Params (M) & GFLOPS & Activations & Memory (M) \\ \hline
CARN~\cite{ahn2018fast}   & -                & 72.41     & 1.59      & 130.04 & 232.98      & 2192.74   \\
EDSR~\cite{lim2017enhanced}   & -                 & 72.36     & 1.52      & 130.05 & 232.98      & 2192.74   \\
IMDN~\cite{hui2019lightweight}   & \textcolor{red}{29.13}            & 38.29     & 0.72      & 46.59  & 122.68      & 922.15    \\
RFDN~\cite{liu2020residual}   & \textcolor{blue}{29.04}         & 29.19     & 0.43      & 27.10  & 112.03      & 813.06    \\
EFDN~\cite{wang2022edge}   & 29.01             & 22.12     & \textcolor{blue}{0.27}      & \textcolor{blue}{16.73}  & 111.11      & 687.44    \\
RLFN~\cite{kong2022residual}   & 29.00            & \textcolor{blue}{19.02}     & 0.32      & 19.70  & \textcolor{blue}{80.04}       & \textcolor{red}{494.80}    \\
DIPNet (Ours)   & 29.00         & \textcolor{red}{14.77}     & \textcolor{red}{0.24}      & \textcolor{red}{14.90}  & \textcolor{red}{72.97}       & \textcolor{blue}{521.02}    \\ \hline
\end{tabular}}
\caption{Comparison of our method and some recent efficient super-resolution methods. Times represent the average inference time measured on the DIV2K dataset with an NVIDIA 2080ti in milliseconds (ms). GFLOPS and memory is measured when the input is 256$\times$256. PSNR is the result of testing on DIV2K. The best and second-best results are marked in \textcolor{red}{red} and  \textcolor{blue}{blue} colors, respectively.} 
\label{tab:ntileres}
\end{table*}

\section{Experiments}

\subsection{Settings and Details}
\textbf{Training and Test Datasets.}
We adopt widely used high-quality (2K resolution) DIV2K~\cite{agustsson2017ntire} dataset which includes 800 training samples for training, following some of the prior research~\cite{ahn2018fast,hui2019lightweight,liu2020residual,kong2022residual}. We test the performance of our method on five benchmark dataset: Set5, Set14, BSD100, Urban100 and Manga109. We also test the data for DIV2K and LDSIR~\cite{lilsdir}, which are  provided by NTIRE 2023 Challenge on Efficient Super Resolution.

\textbf{Evaluation and metrics.}
We use two common metrics called peak signal-to-noise ratio (PSNR) and structure similarity index (SSIM) to evaluate our method and comparison methods on the RGB space, following the evaluation settings of the NTIRE 2023 challenge on efficient super-resolution. In addition, to verify the efficiency of our methods, we statistics the number of parameters, GFLOPs, and inference time of each method network structure in a standard way as validation metrics, which are obtained statistically in the same computing environment.

\textbf{Comparison methods.}
We select representative open-sourced methods, which include CARN~\cite{ahn2018fast}, IMDN~\cite{hui2019lightweight}, RFDN~\cite{liu2020residual}, RLFN~\cite{kong2022residual} and so on. The results of each method are generated by the implementations from the original authors with default settings for a fair comparison.

\textbf{Implementation details.}
All training experiments are done on NVIDIA 2080ti. During the training phase, we use random flip and rotation augmentation and  choose Adam as the optimizer. When training the teacher net and distilling the student network, we set an initial learning rate of 1e-4, halved the learning rate every 100,000 iterations, and then used \(L_1\) loss for supervision. When using the progressive learning strategy to finetune the distilled network, an initial learning rate of 2e-5 is used, which is halved every 20,000 iterations. In this process, the training patch size is progressively increased to improve the performance, which is selected from \([64, 128, 256, 384]\). In the iterative pruning process, the ratio of each pruning is 0.05, and it is repeated three times in total. After each pruning, \(I_{LR}\) and \(I_{enh}\) are used for finetune, and \(384\times384\) patches are used as input during finetune.

\subsection{Model Compexity}
In Fig.~\ref{fig:teaser}, we provide an overview of the deployment performance of our DIPNet. We can find that our DIPNet obtains the best inference time. To evaluate the method complexity of our model precisely, we compare several representative open source networks in Table~\ref{tab:ntileres}. The table shows that our DIPNet consumes the least resource while maintaining 29.00 PSNR. Specifically, in terms of running time, we compare our approach with RFDN on an NVIDIA 2080ti. The iterative pruning strategy that we employ enables significant speed improvements with minimal cost, our speed is significantly faster than RFDN.
\begin{table*}[]
\centering
\setlength{\tabcolsep}{1.2mm}{
\begin{tabular}{cccccccc}
\hline
\multirow{2}{*}{Scale} & \multirow{2}{*}{Model}   & Params                & Set5                               & Set14                              & BSD100                             & UrBan100                           & Manga109                           \\
                       &                          &  (M)                        & PSNR/SSIM                          & PSNR/SSIM                          & PSNR/SSIM                          & PSNR/SSIM                          & PSNR/SSIM                          \\ \hline
\multirow{14}{*}{$\times$2}   & SRCNN~\cite{simonyan2014very}                    & 0.024                       & 36.66 / 0.9542                      & 32.42 / 0.9063                     & 31.36 / 0.8879                     & 29.50 / 0.8946                     & 35.60 / 0.9663                       \\
                       & FSRCNN~\cite{dong2016accelerating}                   & 0.012                       & 36.98 / 0.9556                     & 32.62 / 0.9087                     & 31.50 / 0.8904                     & 29.85 / 0.9009                     & 36.67 / 0.9710                       \\
                       & VDSR~\cite{kim2016accurate}                     & 0.666                      & 37.53 / 0.9587                     & 33.05 / 0.9127                     & 31.90 / 0.8960                     & 30.77 / 0.9141                     & 37.22 / 0.9750                       \\    
                       & DRCN~\cite{kim2016deeply}                     & 1.774                     & 37.63 / 0.9588                     & 33.04 / 0.9118                     & 31.85 / 0.8942                     & 30.75 / 0.9133                     & 37.55 / 0.9732                       \\
                       & LapSRN~\cite{lai2017deep}                   & 0.251                      & 37.52 / 0.9591                     & 32.99 / 0.9124                     & 31.80 / 0.8952                     & 30.41 / 0.9103                     & 37.27 / 0.9740                     \\
                       & IDN~\cite{hui2018fast}                      & 0.579                      & 37.83 / 0.9600                     & 33.30 / 0.9148                     & 32.08 / 0.8985                     & 31.27 / 0.9196                     & 38.01 / 0.9749                     \\
                       & EDSR~\cite{lim2017enhanced}                     & 1.370                     & 37.91 / 0.9602                     & 33.53 / 0.9172                     & 32.15 / 0.8995                     & 31.99 / 0.9270                     & 38.40 / 0.9766                     \\
                       & CARN~\cite{ahn2018fast}                     & 1.592                     & 37.76 / 0.9590                     & 33.52 / 0.9166                     & 32.09 / 0.8978                     & 31.92 / 0.9256                     & 38.36 / 0.9765                     \\
                       
                       & ECBSR~\cite{zhang2021edge}                    & 0.596                      & 37.90 / \textcolor{red}{0.9615}                     & 33.34 / 0.9178                     & 32.10 / \textcolor{red}{0.9018}                     & 31.71 / 0.9250                     & - / -                                \\
                       & IMDN~\cite{hui2019lightweight}                     & 0.694                      & \textcolor{blue}{38.00} / 0.9605                     & 33.63 / 0.9177                     & \textcolor{blue}{32.19} / 0.8996                     & \textcolor{blue}{32.17} / \textcolor{blue}{0.9283}                     & \textcolor{red}{38.88} / \textcolor{red}{0.9774}                     \\
                       & RFDN~\cite{liu2020residual}                     & 0.534                      & \textcolor{red}{38.05} / \textcolor{blue}{0.9606}                     & \textcolor{red}{33.68} / \textcolor{blue}{0.9184}                     & 32.16 / 0.8994                     & 32.12 / 0.9278                     & \textcolor{blue}{38.88} / \textcolor{blue}{0.9773}                     \\
                       & Ours                     & 0.527                      & 37.98 / 0.9605                       & \textcolor{blue}{33.66} / \textcolor{red}{0.9192}                       & \textcolor{red}{32.20} / \textcolor{blue}{0.9002}                       & \textcolor{red}{32.31} / \textcolor{red}{0.9302}                       & 38.62 / 0.9770                       \\ \hline
\multirow{14}{*}{$\times$4}   & SRCNN~\cite{simonyan2014very}                    & 0.057                       & 30.48 / 0.8628                       & 27.49 / 0.7503                       & 26.90 / 0.7101                       & 24.52 / 0.7221                     & 27.58 / 0.8555                       \\
                       & FSRCNN~\cite{dong2016accelerating}                   & 0.013                       & 30.72 / 0.8660                       & 27.61 / 0.7550                       & 26.98 / 0.7150                       & 24.62 / 0.7280                     & 27.90 / 0.8610                       \\
                       & VDSR~\cite{kim2016accurate}                      & 0.666                      & 31.35 / 0.8838                       & 28.01 / 0.7674                       & 27.29 / 0.7251                       & 25.18 / 0.7524                     & 28.83 / 0.8870                       \\
                       & DRCN~\cite{kim2016deeply}                     & 1.774                     & 31.53 / 0.8854                       & 28.02 / 0.7670                       & 27.23 / 0.7233                       & 25.14 / 0.7510                     & 28.93 / 0.8854                       \\
                       & LapSRN~\cite{lai2017deep}                   & 0.502                      & 31.54 / 0.8852                       & 28.09 / 0.7700                       & 27.32 / 0.7275                       & 25.21 / 0.7562                     & 29.09 / 0.8900   \\
                       & IDN~\cite{hui2018fast}   & 0.600  & 31.93 / 0.8923 & 28.45 / 0.7781 & 27.48 / 0.7326 & 25.81 / 0.7766 & 30.04 / 0.9026 \\
                       &  EDSR~\cite{lim2017enhanced} & 1.518 & 31.98 / 0.8927 & 28.55 / 0.7805 & 27.54 / 0.7348 & 25.90 / 0.7809 & 30.24 / 0.9053 \\
                       & CARN~\cite{ahn2018fast}                     & 1.592                     & 32.13 / 0.8937                       & \textcolor{blue}{28.60} / 0.7806                       & \textcolor{blue}{27.58} / 0.7349                       & 26.07 / 0.7837                     & 30.47 / 0.9084                     \\

                       & ECBSR~\cite{zhang2021edge}                     & 0.603                      & 31.92 / 0.8946                       & 28.34 / \textcolor{blue}{0.7817}                       & 27.48 / \textcolor{red}{0.7393}                       & 25.81 
 / 0.7773                      & - / -                                \\
                       & IMDN~\cite{hui2019lightweight}                     & 0.715                      & \textcolor{blue}{32.21} / 0.8948                       & 28.58 / 0.7811                       & 27.56 / 0.7353                       & 26.04 / 0.7838                     & 30.45 / 0.9075 \\
                       & RFDN~\cite{liu2020residual}                     & 0.550                      & \textcolor{red}{32.24} / \textcolor{red}{0.8952}                      & \textcolor{red}{28.61} / \textcolor{red}{0.7819}                       & 27.57 / 0.7360                       & \textcolor{blue}{26.11} / \textcolor{blue}{0.7858}                     & \textcolor{red}{30.58} / \textcolor{red}{0.9089}                     \\
                       & Ours                     & 0.543                      & 32.20 / \textcolor{blue}{0.8950}                       & 28.58 / 0.7811                       & \textcolor{red}{27.59} / \textcolor{blue}{0.7364}                       & \textcolor{red}{26.16} / \textcolor{red}{0.7879}                       & \textcolor{blue}{30.53} / \textcolor{blue}{0.9087}                       \\ \hline
\end{tabular}}
\caption{Quantitative results of the state-of-the-art efficient super-resolution models on four benchmark datasets. The best and second-best results are marked in \textcolor{red}{red} and  \textcolor{blue}{blue} colors, respectively.}
\label{tab:quantitative}
\end{table*}

\subsection{Qualitative Comparison}
As shown in Fig.~\ref{fig:resvis}, we compare our method with some recent efficient super-resolution methods. As can be seen from the figure, although our model is smaller, we can still obtain a good super-resolution effect. Compared with other larger methods, there is no obvious difference in the super-resolution effect. Even in some scenarios, the super-resolution effect of our method is more obvious, such as in the first and second rows in the figure, our method gets clearer lines. This is due to our use of enhanced ground-truth (GT), which makes our method more inclined to learn clearer objects during training.
\begin{table}[]
\centering
\setlength{\tabcolsep}{6.9mm}{
\begin{tabular}{ccc}
\hline
GT Type & \(\mathcal{T}\)    & \(\mathcal{S}\)    \\ \hline
Ori.    & 31.27 & 28.94 \\
Enh.    & 31.20 & 29.00 \\ \hline
\end{tabular}}
\caption{PSNR of the teacher network and the student network on the DIV2K validation set when using the original GT and the enhanced GT, respectively. Ori. means training with original GT, and Enh. means training with enhanced GT.}
\label{tab:enh}
\end{table}
\subsection{Quantitative Comparison}
As shown in Table~\ref{tab:quantitative}, we compare our method with some other state-of-the-art efficient super-resolution models on four branchmark datasets Set5, Set14, and UrBan100. Experiments show that our method still achieves good results on these datasets. It is worth noting that here we do not directly use the final model used in the NTIRE competition in order to achieve better results, but used a larger model with the similar structure, but even so our model is still relatively small.

In Fig.~\ref{fig:vis2} we show some examples of our method and other methods on these datasets. It can be found that our method is much better than other methods in some densely textured areas. This is due to the fact that we use enhanced GT to make our method can better distinguish these repeated regular contents.
\begin{table}[]
\centering
\setlength{\tabcolsep}{2.9mm}{
\begin{tabular}{ccc}
\hline
Shallow Features KD & Deep Feature & PSNR  \\ \hline
                   &             & 28.95 \\
\usym{1F5F8}                   &             & 28.96 \\
                  & \usym{1F5F8}            & 28.98 \\
\usym{1F5F8}                  &\usym{1F5F8}           & 29.00 \\ \hline
\end{tabular}}
\caption{Effect of using different degrees of distillation on PSNR. KD means knowledge distillation, Deep Feature means the features of the last block, and Shallow Features means the features of the first three blocks.} 
\label{tab:KD}
\end{table}
\begin{table*}[!h]
\centering
\setlength{\tabcolsep}{2.9mm}{
\begin{tabular}{ccccccc}
\hline
Method & PSNR  & Time (ms) & Params (M) & GFLOPS & Activations & Memory (M) \\ \hline
KaiBai\_Group  & 28.95                 & 20.49     & 0.272      & 16.76 & 65.10      & 296.45   \\
Young  & 28.97                 & 22.09     & 0.543      & 33.38 & \textcolor{red}{61.87}      & 293.05   \\
NoahTerminalCV\_TeamB  & 28.96                 & 27.83     & 0.209      & \textcolor{red}{13.34} & 118.71      & 188.21   \\
Sissie\_Lab  & 29.00                 & 30.34     & 0.461      & 28.85 & 107.07      & 628.94   \\

Antins\_cv   & 29.00            & 20.92     & 0.315      & 20.07  & 70.82       & 488.61    \\

CMVG   & \textcolor{blue}{29.01}            & 24.42     & 0.307      & 18.98  & 81.55      & 454.51   \\
DFCDN   & 29.00            & 18.71     & \textcolor{blue}{0.245}      & 15.49  & 82.76       & \textcolor{red}{376.99}    \\
Zapdos   & 28.96                & \textcolor{blue}{18.59}     & 0.352      & 21.97 & \textcolor{blue}{63.01}      & \textcolor{blue}{420.50}   \\
DIPNet (Ours)   & \textcolor{red}{29.04}         & \textcolor{red}{18.30}     & \textcolor{red}{0.243}      & \textcolor{blue}{14.90}  & 72.97       & 495.91    \\ \hline
\end{tabular}}
\caption{Quantitative comparison of our results with those of other NTIRE 2023 Challenge on Efficient Super Resolution participating teams. The best and second-best results are marked in \textcolor{red}{red} and  \textcolor{blue}{blue} colors, respectively.} 
\label{tab:esr}
\end{table*}
\begin{table}[]
\centering
\setlength{\tabcolsep}{1.9mm}{
\begin{tabular}{ccccc}
\hline
Pruning times & 0      & 1                          & 2                          & 3                          \\ \hline
PSNR          & 29.042 & \multicolumn{1}{c}{29.034} & \multicolumn{1}{c}{29.018} & \multicolumn{1}{c}{29.001} \\ \hline
\end{tabular}}
\caption{In the iterative pruning process, the PSNR of the model on the DIV2K validation set after each pruning. The \(0\)th represents the PSNR of the model before pruning.}
\label{tab:difftimes}
\end{table}
\begin{table}[]
\centering
\setlength{\tabcolsep}{1.9mm}{
\begin{tabular}{ccc}
\hline
Pruning Type & One Stage Pruning& Iterative pruning  \\ \hline
L1           &  28.883          & 28.946             \\
L2           &   28.894           & 29.001            \\ \hline
\end{tabular}}
\caption{The PSNR when the model is cut to the same size using different pruning strategies}
\label{tab:difftypePrue}
\end{table}
\subsection{Ablation Studies}
\textbf{Original GT vs. Enhanced GT. }
We compared our enhanced ground-truth with the original ground-truth in Fig.~\ref{fig:enhanceGT}, and we can see that the enhanced ground-truth has a clearer texture. And this enhanced ground-truth will help our student network to better learn weaker textures. We further make a quantitative analysis of the effect of enhanced ground-truth in Table~\ref{tab:enh}. From the table, we find that the use of enhanced ground-truth for the teacher network makes the performance of the network worse. This is because the large network has a strong learning ability. It already has the ability to learn most kinds of details, and the use of enhanced ground-truth makes it learn more noise, which leads to a decrease in its PSNR. The student network has a limited learning capability that restricts its ability to explore all possible directions during the training process. As a result, the network may only focus on certain directions and fail to learn other important features necessary for high-quality image generation. However, by incorporating an enhanced ground-truth, which contains more information than the original ground-truth, the student network can overcome its limited capacity to learn details and achieve better performance in terms of PSNR. The enhanced ground-truth provides additional guidance to the network, allowing it to learn a wider range of features and produce more accurate and detailed images.

\textbf{Multi-stage feature distillation.}
The results in Table~\ref{tab:KD} show the advantages of our multi-level feature distillation. Compared with direct end-to-end training of small models, our method can significantly improve the model accuracy by about 0.05dB, which benefits from the strong representation of the teacher model capacity. At the same time, our student model does not add additional overhead. Meanwhile, we found that only using deep features is better than using only shallow features, and the effect is best when the two are used at the same time.

\textbf{Iterative Pruning Strategy.}
We show the changes of PSNR after each pruning in the iterative pruning process in Table~\ref{tab:difftimes}. It can be seen that the PSNR decline in the previous pruning is relatively small, and there will be a relatively obvious decline in PSNR when pruning later. But overall, the reduction in PSNR brought by our iterative pruning method is very small. Table~\ref{tab:difftypePrue} shows the impact of using different pruning methods on the results. The experimental results show that using iterative pruning is better than one-time pruning. At the same time, L2 pruning is better than L1 pruning for our task.
\subsection{NTIRE 2023 Challenge on Efficient SR}
In this competition, we design a lightweight network to maintain the PSNR of 29.00dB on DIV2K validation set by reparameterization and multi-stage feature distillation. It is worth mentioning that the baseline of the competition is RFDN.  We follow the official evaluation setting and report the number of parameters, FLOPs, runtime, peak memory consumption, activations, and number of convolutions in Table~\ref{tab:esr}. Compared with AIM 2020 winner solution E-RFDN, our model can decrease 43.9$\%$ parameters, 45.0$\%$ FLOPs, 48.5$\%$ runtime, 37.1$\%$ peak memory consumption and 34.9$\%$ activations. Compared with other participants in NTIRE 2023 challenge~\cite{li2023ntire_esr} on efficient super-resolution, our model achieves the best inference speed.

\section{Conclusion}
In this paper, we propose a novel approach to efficient single image super-resolution by improving training strategies instead of solely depending on network design. Specifically, we leverage enhanced ground-truth images as additional supervision and employ a multi-stage lightweight training strategy that combines distillation, progressive learning, and pruning. Our experiments demonstrate the effectiveness of our method, achieving state-of-the-art performance in terms of time consumption and model size on the NTIRE 2023 efficient super-resolution challenge. Our contributions include introducing the use of enhanced GT images to improve the learning ability of lightweight networks and proposing a novel multi-stage lightweight training strategy.

{\small
\bibliographystyle{unsrt}
\bibliography{egbib}

\begin{thebibliography}{10}

\bibitem{chen2018fsrnet}
Yu~Chen, Ying Tai, Xiaoming Liu, Chunhua Shen, and Jian Yang.
\newblock Fsrnet: End-to-end learning face super-resolution with facial priors.
\newblock In {\em CVPR}, pages 2492--2501, 2018.

\bibitem{fu2019hyperspectral}
Ying Fu, Tao Zhang, Yinqiang Zheng, Debing Zhang, and Hua Huang.
\newblock Hyperspectral image super-resolution with optimized rgb guidance.
\newblock In {\em CVPR}, pages 11661--11670, 2019.

\bibitem{dong2014learning}
Chao Dong, Chen~Change Loy, Kaiming He, and Xiaoou Tang.
\newblock Learning a deep convolutional network for image super-resolution.
\newblock In {\em ECCV}, pages 184--199, 2014.

\bibitem{hui2019lightweight}
Zheng Hui, Xinbo Gao, Yunchu Yang, and Xiumei Wang.
\newblock Lightweight image super-resolution with information
  multi-distillation network.
\newblock In {\em ACM MM}, pages 2024--2032, 2019.

\bibitem{ignatov2021real}
Andrey Ignatov, Radu Timofte, Maurizio Denna, and Abdel Younes.
\newblock Real-time quantized image super-resolution on mobile npus, mobile ai
  2021 challenge: Report.
\newblock In {\em CVPR}, pages 2525--2534, 2021.

\bibitem{tai2017image}
Ying Tai, Jian Yang, and Xiaoming Liu.
\newblock Image super-resolution via deep recursive residual network.
\newblock In {\em CVPR}, pages 3147--3155, 2017.

\bibitem{shi2016real}
Wenzhe Shi, Jose Caballero, Ferenc Husz{\'a}r, Johannes Totz, Andrew~P Aitken,
  Rob Bishop, Daniel Rueckert, and Zehan Wang.
\newblock Real-time single image and video super-resolution using an efficient
  sub-pixel convolutional neural network.
\newblock In {\em CVPR}, pages 1874--1883, 2016.

\bibitem{liu2020residual}
Jie Liu, Jie Tang, and Gangshan Wu.
\newblock Residual feature distillation network for lightweight image
  super-resolution.
\newblock In {\em ECCV}, pages 41--55, 2020.

\bibitem{wei2020aim}
Pengxu Wei, Hannan Lu, Radu Timofte, Liang Lin, Wangmeng Zuo, Zhihong Pan,
  Baopu Li, Teng Xi, Yanwen Fan, Gang Zhang, et~al.
\newblock Aim 2020 challenge on real image super-resolution: Methods and
  results.
\newblock pages 392--422, 2020.

\bibitem{zhang2021edge}
Xindong Zhang, Hui Zeng, and Lei Zhang.
\newblock Edge-oriented convolution block for real-time super resolution on
  mobile devices.
\newblock In {\em ACM MM}, 2021.

\bibitem{ahn2018fast}
Namhyuk Ahn, Byungkon Kang, and Kyung-Ah Sohn.
\newblock Fast, accurate, and lightweight super-resolution with cascading
  residual network.
\newblock In {\em ECCV}, 2018.

\bibitem{hui2018fast}
Zheng Hui, Xiumei Wang, and Xinbo Gao.
\newblock Fast and accurate single image super-resolution via information
  distillation network.
\newblock In {\em CVPR}, pages 723--731, 2018.

\bibitem{chen2023human}
Du~Chen, Jie Liang, Xindong Zhang, Ming Liu, Hui Zeng, and Lei Zhang.
\newblock Human guided ground-truth generation for realistic image
  super-resolution.
\newblock In {\em CVPR}, 2023.

\bibitem{ding2021repvgg}
Xiaohan Ding, Xiangyu Zhang, Ningning Ma, Jungong Han, Guiguang Ding, and Jian
  Sun.
\newblock Repvgg: Making vgg-style convnets great again.
\newblock In {\em CVPR}, pages 13733--13742, 2021.

\bibitem{chu2022improving}
Xiaojie Chu, Liangyu Chen, Chengpeng Chen, and Xin Lu.
\newblock Improving image restoration by revisiting global information
  aggregation.
\newblock In {\em ECCV}, pages 53--71, 2022.

\bibitem{simonyan2014very}
Karen Simonyan and Andrew Zisserman.
\newblock Very deep convolutional networks for large-scale image recognition.
\newblock {\em arXiv preprint arXiv:1409.1556}, 2014.

\bibitem{kim2016accurate}
Jiwon Kim, Jung~Kwon Lee, and Kyoung~Mu Lee.
\newblock Accurate image super-resolution using very deep convolutional
  networks.
\newblock In {\em CVPR}, pages 1646--1654, 2016.

\bibitem{lim2017enhanced}
Bee Lim, Sanghyun Son, Heewon Kim, Seungjun Nah, and Kyoung Mu~Lee.
\newblock Enhanced deep residual networks for single image super-resolution.
\newblock In {\em CVPRW}, pages 136--144, 2017.

\bibitem{zhang2018residual}
Yulun Zhang, Yapeng Tian, Yu~Kong, Bineng Zhong, and Yun Fu.
\newblock Residual dense network for image super-resolution.
\newblock In {\em CVPR}, pages 2472--2481, 2018.

\bibitem{haris2018deep}
Muhammad Haris, Gregory Shakhnarovich, and Norimichi Ukita.
\newblock Deep back-projection networks for super-resolution.
\newblock In {\em CVPR}, pages 1664--1673, 2018.

\bibitem{luo2021ebsr}
Ziwei Luo, Lei Yu, Xuan Mo, Youwei Li, Lanpeng Jia, Haoqiang Fan, Jian Sun, and
  Shuaicheng Liu.
\newblock Ebsr: Feature enhanced burst super-resolution with deformable
  alignment.
\newblock In {\em CVPR}, pages 471--478, 2021.

\bibitem{zhang2018image}
Yulun Zhang, Kunpeng Li, Kai Li, Lichen Wang, Bineng Zhong, and Yun Fu.
\newblock Image super-resolution using very deep residual channel attention
  networks.
\newblock In {\em ECCV}, 2018.

\bibitem{zhang2020residual}
Yulun Zhang, Yapeng Tian, Yu~Kong, Bineng Zhong, and Yun Fu.
\newblock Residual dense network for image restoration.
\newblock {\em IEEE TPAMI}, 43(7):2480--2495, 2020.

\bibitem{luo2022bsrt}
Ziwei Luo, Youwei Li, Shen Cheng, Lei Yu, Qi~Wu, Zhihong Wen, Haoqiang Fan,
  Jian Sun, and Shuaicheng Liu.
\newblock Bsrt: Improving burst super-resolution with swin transformer and
  flow-guided deformable alignment.
\newblock In {\em CVPR}, pages 998--1008, 2022.

\bibitem{ledig2017photo}
Christian Ledig, Lucas Theis, Ferenc Husz{\'a}r, Jose Caballero, Andrew
  Cunningham, Alejandro Acosta, Andrew Aitken, Alykhan Tejani, Johannes Totz,
  Zehan Wang, et~al.
\newblock Photo-realistic single image super-resolution using a generative
  adversarial network.
\newblock In {\em CVPR}, pages 4681--4690, 2017.

\bibitem{sajjadi2017enhancenet}
Mehdi~SM Sajjadi, Bernhard Scholkopf, and Michael Hirsch.
\newblock Enhancenet: Single image super-resolution through automated texture
  synthesis.
\newblock In {\em ICCV}, pages 4491--4500, 2017.

\bibitem{wang2018esrgan}
Xintao Wang, Ke~Yu, Shixiang Wu, Jinjin Gu, Yihao Liu, Chao Dong, Yu~Qiao, and
  Chen Change~Loy.
\newblock Esrgan: Enhanced super-resolution generative adversarial networks.
\newblock In {\em ECCV}, pages 0--0, 2018.

\bibitem{Ji_2020_CVPR_Workshops}
Xiaozhong Ji, Yun Cao, Ying Tai, Chengjie Wang, Jilin Li, and Feiyue Huang.
\newblock Real-world super-resolution via kernel estimation and noise
  injection.
\newblock In {\em CVPRW}, 2020.

\bibitem{li2022d2c}
Youwei Li, Haibin Huang, Lanpeng Jia, Haoqiang Fan, and Shuaicheng Liu.
\newblock D2c-sr: A divergence to convergence approach for real-world image
  super-resolution.
\newblock In {\em ECCV}, pages 379--394, 2022.

\bibitem{gu2019blind}
Jinjin Gu, Hannan Lu, Wangmeng Zuo, and Chao Dong.
\newblock Blind super-resolution with iterative kernel correction.
\newblock In {\em CVPR}, pages 1604--1613, 2019.

\bibitem{bell2019blind}
Sefi Bell-Kligler, Assaf Shocher, and Michal Irani.
\newblock Blind super-resolution kernel estimation using an internal-gan.
\newblock {\em NeurIPS}, 32, 2019.

\bibitem{luo2022deep}
Ziwei Luo, Haibin Huang, Lei Yu, Youwei Li, Haoqiang Fan, and Shuaicheng Liu.
\newblock Deep constrained least squares for blind image super-resolution.
\newblock In {\em CVPR}, pages 17642--17652, 2022.

\bibitem{johnson2016perceptual}
Justin Johnson, Alexandre Alahi, and Li~Fei-Fei.
\newblock Perceptual losses for real-time style transfer and super-resolution.
\newblock In {\em ECCV}, pages 694--711, 2016.

\bibitem{goodfellow2014generative}
Ian~J Goodfellow, Jean Pouget-Abadie, Mehdi Mirza, Bing Xu, David Warde-Farley,
  Sherjil Ozair, Aaron Courville, and Yoshua Bengio.
\newblock Generative adversarial networks.
\newblock {\em arXiv preprint arXiv:1406.2661}, 2014.

\bibitem{liang2021swinir}
Jingyun Liang, Jiezhang Cao, Guolei Sun, Kai Zhang, Luc Van~Gool, and Radu
  Timofte.
\newblock Swinir: Image restoration using swin transformer.
\newblock In {\em ICCV}, pages 1833--1844, 2021.

\bibitem{chen2022activating}
Xiangyu Chen, Xintao Wang, Jiantao Zhou, and Chao Dong.
\newblock Activating more pixels in image super-resolution transformer.
\newblock {\em arXiv preprint arXiv:2205.04437}, 2022.

\bibitem{kong2022residual}
Fangyuan Kong, Mingxi Li, Songwei Liu, Ding Liu, Jingwen He, Yang Bai, Fangmin
  Chen, and Lean Fu.
\newblock Residual local feature network for efficient super-resolution.
\newblock In {\em CVPR}, pages 766--776, 2022.

\bibitem{du2022fast}
Zongcai Du, Ding Liu, Jie Liu, Jie Tang, Gangshan Wu, and Lean Fu.
\newblock Fast and memory-efficient network towards efficient image
  super-resolution.
\newblock In {\em CVPR}, pages 853--862, 2022.

\bibitem{luo2023fast}
Ziwei Luo, Youwei Li, Lei Yu, Qi~Wu, Zhihong Wen, Haoqiang Fan, and Shuaicheng
  Liu.
\newblock Fast nearest convolution for real-time efficient image
  super-resolution.
\newblock pages 561--572, 2023.

\bibitem{li2022ntire}
Yawei Li, Kai Zhang, Radu Timofte, Luc Van~Gool, Fangyuan Kong, Mingxi Li,
  Songwei Liu, Zongcai Du, Ding Liu, Chenhui Zhou, et~al.
\newblock Ntire 2022 challenge on efficient super-resolution: Methods and
  results.
\newblock In {\em CVPR}, pages 1062--1102, 2022.

\bibitem{ignatov2023efficient}
Andrey Ignatov, Radu Timofte, Maurizio Denna, Abdel Younes, Ganzorig Gankhuyag,
  Jingang Huh, Myeong~Kyun Kim, Kihwan Yoon, Hyeon-Cheol Moon, Seungho Lee,
  et~al.
\newblock Efficient and accurate quantized image super-resolution on mobile
  npus, mobile ai \& aim 2022 challenge: report.
\newblock pages 92--129, 2023.

\bibitem{zhu2017prune}
Michael Zhu and Suyog Gupta.
\newblock To prune, or not to prune: exploring the efficacy of pruning for
  model compression.
\newblock {\em arXiv preprint arXiv:1710.01878}, 2017.

\bibitem{wang2022edge}
Yan Wang.
\newblock Edge-enhanced feature distillation network for efficient
  super-resolution.
\newblock In {\em CVPR}, pages 777--785, 2022.

\bibitem{agustsson2017ntire}
Eirikur Agustsson and Radu Timofte.
\newblock {NTIRE} 2017 challenge on single image super-resolution: Dataset and
  study.
\newblock In {\em CVPRW}, pages 126--135, 2017.

\bibitem{lilsdir}
Yawei Li, Kai Zhang, Jingyun Liang, Jiezhang Cao, Ce~Liu, Rui Gong, Yulun
  Zhang, Hao Tang, Yun Liu, Denis Demandolx, Rakesh Ranjan, Radu Timofte, and
  Luc Van~Gool.
\newblock Lsdir: A large scale dataset for image restoration.
\newblock In {\em CVPRW}, 2023.

\bibitem{dong2016accelerating}
Chao Dong, Chen~Change Loy, and Xiaoou Tang.
\newblock Accelerating the super-resolution convolutional neural network.
\newblock In {\em ECCV}, pages 391--407, 2016.

\bibitem{kim2016deeply}
Jiwon Kim, Jung~Kwon Lee, and Kyoung~Mu Lee.
\newblock Deeply-recursive convolutional network for image super-resolution.
\newblock In {\em CVPR}, pages 1637--1645, 2016.

\bibitem{lai2017deep}
Wei-Sheng Lai, Jia-Bin Huang, Narendra Ahuja, and Ming-Hsuan Yang.
\newblock Deep laplacian pyramid networks for fast and accurate
  super-resolution.
\newblock In {\em CVPR}, pages 624--632, 2017.

\bibitem{li2023ntire_esr}
Yawei Li, Yulun Zhang, Luc Van~Gool, Radu Timofte, et~al.
\newblock Ntire 2023 challenge on efficient super-resolution: Methods and
  results.
\newblock In {\em CVPRW}, 2023.

\end{thebibliography}
}

\end{document}